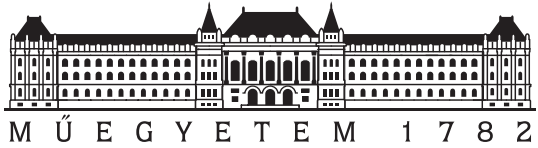

# Improving the sample-efficiency of neural architecture search with reinforcement learning

**Scientific Students' Association Report**

Authors:

Attila Nagy
Ábel Boros

Supervisors:

Hamdi Abed
Dr. Bálint Gyires-Tóth

2019

# Contents

# Kivonat

Az elmúlt évtized deep learning forradalmának egyik legfőbb mozgatórugója az összetett architektúrák megtervezése és implementálása volt. Ma már, amikor adat alapú megközelítéssel igyekszünk megoldani egy bonyolult problémát, bevált módszer egy már jól működő, neves szakértők által megalkotott architektúra (pl. Inception) felhasználása. Ez bizonyos esetekben elegendő, de az igazán összetett, vagy korábban nem ismert típusú feladatok esetén nagy pontosságot csak magasan képzett deep learning szakemberek segítségével lehet elérni.

Munkánkban az automatizált gépi tanulás (Automated Machine Learning, AutoML) területéhez szeretnénk hozzájárulni, azon belül is a neurális architektúrák kereséséhez (Neural Architecture Search, NAS). Ezekkel a módszerekkel bonyolult, probléma-specifikus architektúrák tervezése is lehetséges komoly háttértudás nélkül, ennélfogva a társadalom szélesebb körben tudja kihasználni a mély tanulásban rejlő lehetőségeket. Habár az elmúlt években számos ilyen módszer (pl. gradiens alapú vagy evolúciós algoritmus) jelent meg, dolgozatunkban az egyik legígéretesebb iránnyal, a megerősítéses tanuláson alapuló megközelítéssel foglalkozunk. Ebben az esetben egy rekurrens neurális hálózatot (kontroller) tanítunk probléma-specifikus neurális háló architektúrák generálására. A generált hálózatok validációs pontossága képezi a kontroller megerősítéses tanulással való tanítása során használt jutalom függvényt.

Munkánk alapját képezi az Efficient Neural Architecture Search (ENAS) algoritmus, amelyben súly megosztást alkalmaznak a generált architektúrák között. Az ENAS, mint sok más megerősítéses tanuláson alapuló megközelítés a generált architektúrák konvergenciáján igyekszik javitani. A kontroller így sűrűbb jutalom függvényhez jut, ezáltal lecsökkentve a tanítás idejét. Dolgozatunkban a probléma másik oldalát közelítjük meg, vagyis a kontroller tanításán szeretnénk javítani. A kontroller eredetileg a REINFORCE algoritmus alapján tanult. Kutatómunkánkban mi ezt egy korszerűbb és összetettebb megerősítéses tanulás algoritmusra, a PPO-ra cseréljük le, ami más problémakörökben stabilabb és gyorsabb tanulást eredményezett. Ezután a kapott eredményeket az architektúria keresés aspektusaiból részletesen elemezzük.

# Abstract

Designing complex architectures has been an essential cogwheel in the revolution deep learning has brought about in the past decade. When solving difficult problems in a data-driven manner, a well-tried approach is to take an architecture discovered by renowned deep learning scientists as a basis (e.g. Inception) and try to apply it to a specific problem. This might be sufficient, but as of now, achieving very high accuracy on a complex or yet unsolved task requires the knowledge of highly-trained deep learning experts.

In this work, we would like to contribute to the area of Automated Machine Learning (AutoML), specifically Neural Architecture Search (NAS), which intends to make deep learning methods available for a wider range of society by designing neural topologies automatically. Although several different approaches exist (e.g. gradient-based or evolutionary algorithms), our focus is on one of the most promising research directions, reinforcement learning. In this scenario, a recurrent neural network (controller) is trained to create problem-specific neural network architectures (child). The validation accuracies of the child networks serve as a reward signal for training the controller with reinforcement learning.

The basis of our proposed work is Efficient Neural Architecture Search (ENAS), where parameter sharing is applied among the child networks. ENAS, like many other RL-based algorithms, emphasize the learning of child networks as increasing their convergence result in a denser reward signal for the controller, therefore significantly reducing training times. The controller was originally trained with REINFORCE. In our research, we propose to modify this to a more modern and complex algorithm, PPO, which has demonstrated to be faster and more stable in other environments. Then, we briefly discuss and evaluate our results.

# Introduction

Although the theoretical foundations of machine learning and neural networks have been around for more than 50 years, its' applications have only become widespread in the last decade. This is due to the increasing amount of available computing power, more efficient algorithms, better frameworks for applying machine learning in practice and obviously the enormous amount of data available at hand. If we take a look at the technological trends of recent years, it is beyond dispute that a large number of research groups and companies decided to introduce data-driven solutions. The wide area of machine learning and statistical modelling are what make these processes highly efficient and lucrative when it comes to delivering business value. Designing these algorithms requires highly-trained professionals with excellent application domain expertise. As there is an increasing demand for these engineers and scientists, the idea of automating parts, or even the whole data science pipeline in its entirety has become a topic of frequent discussion. From a social perspective, it is important to address this topic, because by automating machine learning workflows, we enable a wider range of the society to harness the power of machine learning, thus avoiding a future situation, where these methods are available to the few, but not the many. When it comes to designing neural networks for a given task, people often rely on heuristics. From a scientific perspective, the significance of this direction of research not only could be verifying or demystifying these heuristics, but also discovering new patterns in predictive modelling. Automated Machine Learning is essential and holds lots of potential for the industry as well, as the process of developing machine learning algorithms is a long-lasting and costly process of engineering trial-and-error.

In this work, we will cover the theoretical background of neural networks, reinforcement learning and neural architecture search using reinforcement learning. Afterwards, we propose a modified searching algorithm for ENAS using Proximal Policy Optimization, evaluate our results on CIFAR-10, compare the proposed method with the previously used REINFORCE algorithm and argue, that a controller trained with PPO delivers better results in terms of sample-efficiency.

# Chapter 1

# Background

## 1.1 Neural networks

In this section, we will give a summary of neural networks primarily based on the Deep Learning textbook by Ian Goodfellow, Yoshua Bengio and Aaron Courville [10].

### 1.1.1 Feedforward neural networks

A feedforward network is a computational model inspired by biological neural networks. The purpose of such a model is to approximate a function $f$ for a given input vector $\mathbf{x}$ as:

$$\hat{y} = f(\mathbf{x}; \boldsymbol{\theta}),$$

where $\boldsymbol{\theta}$ denotes the learnable parameters of the network. The model is a composition of $n$ functions (layers):

$$f(\boldsymbol{x}) = f^{(n)}(f^{(n-1)}...(f^{(2)}(f^{(1)}(\mathbf{x})))),$$

where the output of the $n-1$th layer serves as the input of the $n$th layer. By definition, feedforward networks restrict feedback connections, such as directed cycles or loops in the computational graph. Consequently the model can be represented as a directed acyclic graph (as shown in figure 1.1).

The simplest architecture where $n = 1$ is called a single-layer perceptron. In this case we are given an input vector $\mathbf{x}$ for which we compute:

$$\hat{y} = \boldsymbol{\theta}\mathbf{x} + b,$$

where $\boldsymbol{\theta}$ are the learnable weights in the singular layer and $b$ denotes the bias. This by far is a linear model, which can only approximate linear functions, therefore in order to extend the set of representable functions, a $\phi$ nonlinearity is added to the model in the form of an activation function:

$$z = \phi(\boldsymbol{\theta}\mathbf{x} + b)$$

The Universal Approximation Theorem [13, 9] states that a feedforward network having a linear output layer and at least one hidden layer squashed through an activation function is capable of approximating even nonlinear functions. This however is challenging in practice as the single layer may have an infeasible size and therefore the algorithm may fail to generalise. In practice, connecting multiple relatively narrow layers turned out to be efficient, because it introduces levels of abstraction to the representation of knowledge [17]. Recent advances in machine learning rely heavily on deep learning: the science of designing neural networks, where the number of layers in the model is considered high.

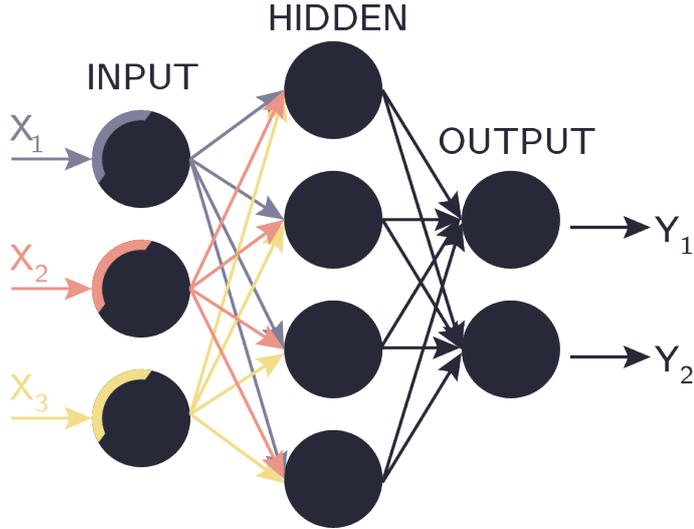

**Figure 1.1:** A neural network as a directed acyclical graph

In order to find the mapping between the input and output of the network, firstly a loss function needs to be defined between the predicted output $\hat{y}$ and the actual output $y$. During the training of a neural network, this is what we intend to minimise by computing the gradient w.r.t. the loss function:

$$\boldsymbol{\theta_{t+1}} \leftarrow \boldsymbol{\theta_t} - \alpha \nabla_{\boldsymbol{\theta}} L(\boldsymbol{\theta}),$$

where $\alpha$ is a positive-definite step size (learning rate) and $L(\boldsymbol{\theta})$ is the defined loss function. This unfortunately applies for one training example, so in order to compute the gradients for a whole iteration through the training data, we need to calculate the gradients for every data point and take the average of the calculated gradients. The data used for training may be huge in size to induce generalization, perhaps consisting of millions of examples, therefore the proposed gradient descent in it's form is computationally very expensive to perform. In practice, an extended version of gradient descent is used, namely stochastic gradient descent (SGD). The motivation behind SGD, is that a gradient can be closely estimated using only a small subset of the training samples, so for every computation of the gradient, a minibatch of examples are sampled uniformly from the training set. This way the gradient computes as:

$$\boldsymbol{g} = \frac{1}{m} \nabla_{\boldsymbol{\theta}} \sum_{i=1}^{m} L(\boldsymbol{\theta}),$$

where $m$ denotes the size of a minibatch. The gradient then computes accordingly as:

$$\boldsymbol{\theta_{t+1}} \leftarrow \boldsymbol{\theta_t} - \alpha \boldsymbol{g}.$$

The parameter updating algorithm can be further improved by using the momentum method to accelerate convergence and avoid situations where the gradient is stuck in local minima by including previously calculated gradients into the computation at the current time step as:

$$v_{t+1} = \gamma v_t - \alpha g$$
$$\theta_{t+1} = \theta_t + v_{t+1},$$

where $v$ is the velocity parameter containing the weighted sum of all previous gradients and $\gamma$ is the decaying factor. Even more frequently used in practice is a modified version of the momentum method, called Nesterov momentum. It introduces one important addition to the previously discussed method, which is related to the calculation of gradients. When calculating $\theta_{t+1}$, instead of taking $\theta_t$ as a starting point in the parameter space, we compute the gradient term from an intermediate point $\theta_{intermediate} = \theta_t + \gamma v_t$. The updated expression for the Nesterov accelerated gradient is:

$$v_{t+1} = \gamma v_t - \alpha \nabla L(\theta_t + \gamma v_t)$$
$$\theta_{t+1} = \theta_t + v_{t+1}$$

The underlying reason for this is that in some cases, the parameter updates might result in a significant increase in the loss function and this information remains encoded in the velocity vector during computations in the future. This results in oscillations, which could significantly slow down the convergence towards the minimum.

As the predicted output is a linear combination of nonlinear terms composed of the input data, the weights and the biases, we actually compute the partial derivatives of the loss function w.r.t. each parameter with backpropagation [26]. This is done via the recursive application of the chain rule.

The selection of activation functions have a huge influence on the convergence of the learning algorithm. When relying on the Universal Approximation Theorem, it is important to note, that it only holds for bounded nonlinearities. These nonlinear functions have a compact range as they squeeze the input values into a bounded subset of the set of real numbers, hence they are often referred to as saturating functions. Traditionally, the functions sigmoid as $\phi_s(x) = \frac{1}{1+e^{-x}}$ (Figure 1.2 left) and hyperbolic tangent as $\phi_{ht}(x) = \frac{e^x - e^{-x}}{e^x + e^{-x}}$ (Figure 1.2 middle) were used in neural networks.

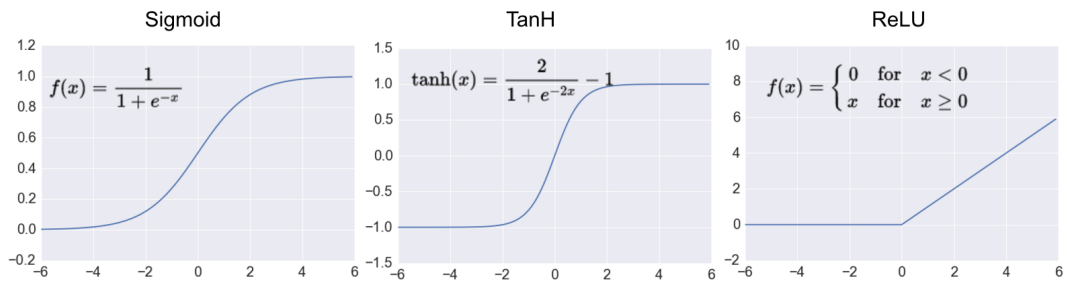

**Figure 1.2:** The three commonly used activation functions, sigmoid (left), hyperbolic tangent (middle) and the rectified linear unit (right) [22]

These however, cause some difficulties in practice. On one hand, the derivative of these functions are close to zero at a significant number of points, which slow down gradient descent. On the other hand, gradients could vanish during optimization, due to the fact that the absolute value produced by these saturating functions is smaller than one. This is especially crucial in deep neural networks, because when the gradient of the error is propagated backwards through the network, the further a parameters is from the output layer, the harder it is to train.

One significant milestone in deep learning research was the introduction of an unbound nonlinearity, namely the rectified linear unit (ReLU) as $\phi_r(x) = max(0, x)$ [23] (Figure 1.2 right). As the positive values are not downscaled during training, ReLU makes the learning algorithm more robust against vanishing gradients. It is worth noting, that ReLU is non-differentiable at $x = 0$, but conventionally it is considered as 0, which fortunately does not cause complications in practice.

### 1.1.2 Convolutional neural networks

Convolutional neural networks are suitable for solving problems involving data with a grid-like structure [16]. As therefore convolutional nets are very suitable for image recognition tasks and we also discuss our methods through image recognition, we will refer to image data while presenting the theory. A convolutional layer applies a special linear operation, where a kernel with weights is sliding through the input data, while multiplying its' weights and the corresponding input pixels. In two-dimensions, performing convolution can be expressed as:

$$S(i,j) = (K * I)(i,j) = \sum_m \sum_n I(i-m, j-n)K(m,n),$$

where $(*)$ denote the convolution operation, $K(m,n)$ is the kernel and $I(i,j)$ is the image. These matrix multiplications can get computationally expensive, therefore it is a good practice to use separable convolutions, by which the number of multiplications can be reduced. This could be spatial or depthwise, but as the latter is used more often in practice, we discuss depthwise separable convolution [8]. Firstly, a depthwise convolution is performed, where the input and filter channels are split and the convolution operations are performed independently (as shown on figure 1.3). The results then are stacked and we perform pointwise convolution with a 1x1 kernel, to achieve the desired output dimension.

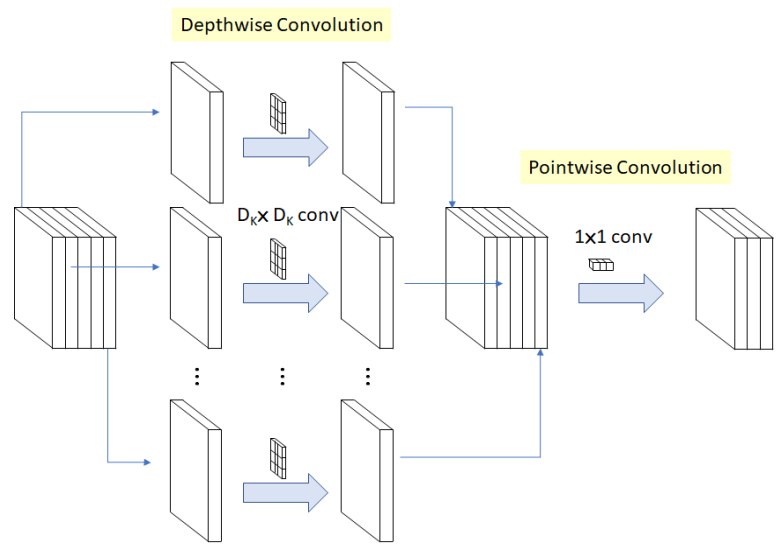

**Figure 1.3:** The two steps of depthwise separable convolution [35]

The next step is to apply an element-wise nonlinearity (usually ReLU) on the feature map created by the convolutional layer. By stacking more convolutional layers, we intend to learn some hierarchical representations in the images, therefore the resolution of the feature map should be reduced while passing through the layers. This reduction can either be done by modifying the step size of the convolution (stride) to a number greater than one or by applying pooling layers. Several types of pooling layers exist, all of them sharing the purpose of reducing model complexity by decreasing the spatial dimensions of the activations. They also make the representation more invariant to small translations of the input image [10]. The two most commonly used pooling layers are average pooling and max pooling. In average pooling, we take a given input block and reduce it to 1x1 by taking the average of the coefficients. In case of max pooling, we similarly perform the operation, but take the maximum of the input values. It is important to note, that pooling does not modify the depth dimension.

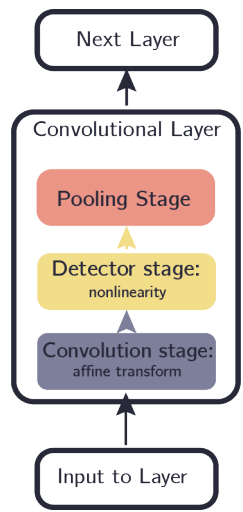

**Figure 1.4:** Components of a convolutional layer

### 1.1.3 Recurrent neural networks

Time is a unique dimension in the context of neural networks as it cannot be treated the same way as any other dimension of the data. Due to the fact that time is strictly progressing in one direction, classical neural networks are unable to efficiently capture the temporal representations in sequential data. However, *Recurrent Neural Networks* [26] were created with the purpose of solving this problem by enabling recurrent connections (loops) in the topology of neural networks. During every iteration, the RNN uses an input vector $\boldsymbol{x_t}$ and a hidden state vector $\boldsymbol{h_{t-1}}$ to compute $\boldsymbol{h_t}$ with the following recursive formula:

$$\boldsymbol{h}_t = f(\boldsymbol{h}_{t-1}, \boldsymbol{x}_t; \boldsymbol{\theta}),$$

where $\boldsymbol{\theta}$ denotes the learnable parameters of the network. As the output of the previous time step serve as the input of the current one, the computational graph can simply be unfolded (as shown on figure 1.5), on which we perform back-propagation through time (BPTT) [37] to compute the gradients during training.

Theoretically, recurrent neural networks are capable of learning dependencies of any length, however it has been shown [6] that in practice the problem of vanishing and exploding gradients make it very difficult to learn long-term dependencies.

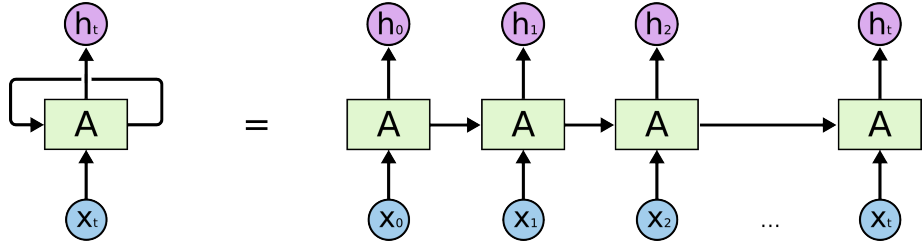

**Figure 1.5:** The unrolled computations in an RNN [24]

#### 1.1.3.1 Long Short-Term Memory networks

A *Long Short-Term Memory Network* (LSTM) [12] is a special kind of RNN capable of overcoming the major shortcoming of simple recurrent neural networks, the problem of long-term dependencies by introducing self-loops. This augmentation prevents exploding/vanishing gradients by enabling them to flow for long durations. Besides the external recurrencies utilized by standard recurrent neural networks, the self-loops of long short-term memory networks are located internally in the LSTM cells. The inputs and outputs of an LSTM and a standard RNN are similar in structure, the crucial difference lies in the more complex internal computations of the LSTM cells. Here, the core component is called the state unit, which is governed by other internal gates, such as the forget gate, input gate and output gate. The forget gate is characterized by the following equation:

$$f_i^{(t)} = \sigma\left(b_i^f + \sum_j U_{i,j}^f x_j^{(t)} + \sum_j W_{i,j}^f h_j^{(t-1)}\right)$$

where $\sigma(\cdot)$ computes an element-wise sigmoid function, $\boldsymbol{x^{(t)}}$ and $\boldsymbol{h^{(t)}}$ denote the input vector and the hidden layer vector at time step $t$. The variables $\boldsymbol{b^f}$, $\boldsymbol{U^f}$ and $\boldsymbol{W^f}$ respectively are the biases, input weights and recurrent weights corresponding to the forget gate. The input gate is structured similarly:

$$g_i^{(t)} = \sigma\left(b_i^g + \sum_j U_{i,j}^g x_j^{(t)} + \sum_j W_{i,j}^g h_j^{(t-1)}\right)$$

having it's own weights and biases respectively denoted by $\boldsymbol{b^g}$, $\boldsymbol{U^g}$ and $\boldsymbol{W^g}$. Governed by the components introduced above, the state unit is denoted with $s_i^{(t)}$ and given by the following equation:

$$s_i^{(t)} = f_i^{(t)} s_i^{(t-1)} + g_i^{(t)} \sigma\left(b_i + \sum_j U_{i,j} x_j^{(t)} + \sum_j W_{i,j} h_j^{(t-1)}\right),$$

where $\boldsymbol{b}, \boldsymbol{U}$ and $\boldsymbol{W}$ respectively mark the biases, input weights and recurrent weights of the LSTM cell. At last, the state unit is passed through a hyperbolic tangent nonlinearity, thus the output is given as:

$$h_i^{(t)} = tanh(s_i^{(t)}) q_i^{(t)},$$

where $q_i^{(t)}$ is responsible for gating the output as:

$$q_i^{(t)} = \sigma\left(b_i + \sum_j U_{i,j} x_j^{(t)} + \sum_j W_{i,j} h_j^{(t-1)}\right),$$

with $\boldsymbol{b}^o$, $\boldsymbol{U}^o$ and $\boldsymbol{W}^o$ denoting the biases, input weights and recurrent weights.

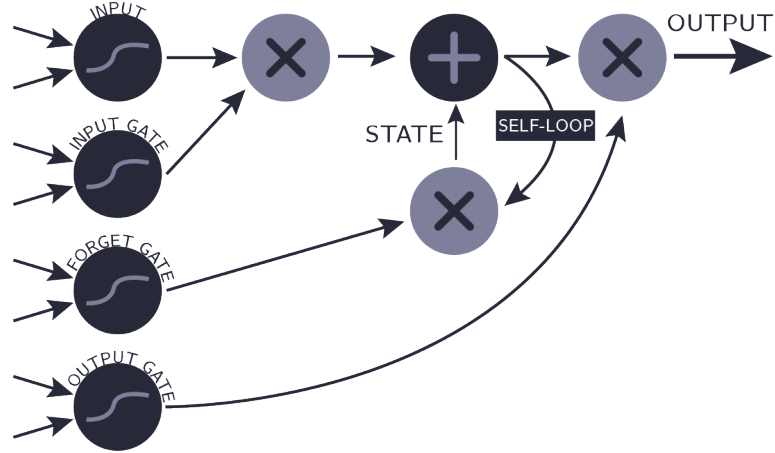

**Figure 1.6:** The internal structure of an LSTM cell

## 1.2 Reinforcement Learning

Reinforcement learning is an area of machine learning, where we aim to solve a decision process by applying some kind of strategy that involves learning from feedback given by an environment. The core of the problem is to develop an intelligent agent that maximizes a predefined reward function by performing a series of actions. Throughout training, the agent makes decisions based on some arbitrary logic and this way interacts with the environment. This interaction has three essential components:

- *Observation:* how the agent sees the current state of the environment,
- *Action:* what the agent performs in light of the observations,
- *Reward:* what the agent receives after performing an action.

Due to the rise of deep learning, novel reinforcement learning methods were developed in the past decade, performing very efficiently in a wide variety of complex environments [7]. For the majority of section 1.2, we rely on Reinforcement Learning: An Introduction by Richard S. Sutton and Andrew G. Barto [32] for presenting the theoretical background of reinforcement learning.

### 1.2.1 Markov Decision Processes

Reinforcement learning problems can be formulated within the mathematical framework of *Markov Decision Processes*. A MDP is characterized by the following 5-tuple:

- **S** state space,
- **A** action space,
- $\mathbf{P}(\mathbf{S_0})$ starting state distribution,
- $\mathbf{P}(\mathbf{s'|s,a})$ transition probability (describing the probability of moving to state $s'$ after performing action $a$ in state $s$),

- **r(s, a, s′)** reward received for performing action $a$ in state $s$ resulting in a transition to state $s'$.

It's name derives from the fact that Markov Decision Processes satisfy the Markov Property, which states that the future is independent of the past given the present. In other words, if a state $S_t$ is already given, there is no need for retaining any of the previous states. Therefore mathematically speaking a state $S_t$ satisfy the Markov Property if and only if:

$$P[S_{t+1}|S_t] = P[S_{t+1}|S_1, S_2, ..., S_t].$$

In a Markov Decision Process, the agent observes the current state of the environment $s_t$ for every time step $t$. The observation, according to what the agent takes an action $a_t$, can either be full or partial. Having performed the action, a transition takes place to state $s_{t+1}$, which can be stochastic or deterministic. The agent then receives a reward $r_t$, which can be zero, positive or negative. The main task of the agent is to maximize the cumulative reward from the very first time step, until the termination of the environment. Since rewards are not necessarily given for every transition and the given results might not reflect the true value of a taken action, it is challenging to estimate the reward corresponding to actions.

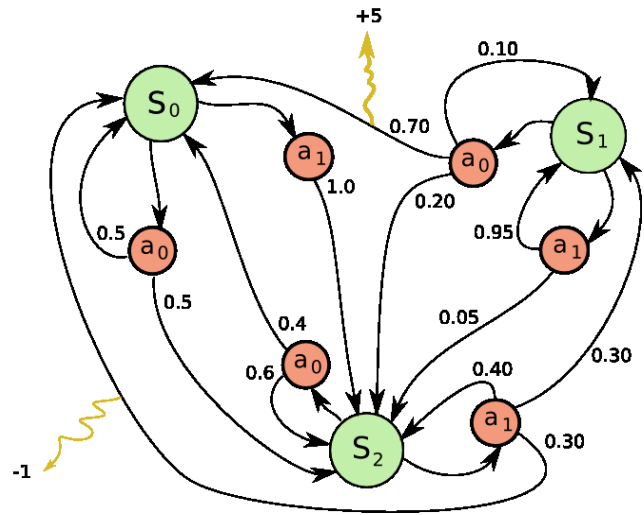

**Figure 1.7:** A simple Markov Decision Process illustrated [4]

### 1.2.2 Preliminaries and notations

#### 1.2.2.1 Policy

A *policy* is a function that associates an action $a$ for every state $s$ as

$$p_i(s) = a.$$

In other words, this is the inner logic that drives the actions of the agent. During training, we want to converge towards the perfect policy $(p_i^*)$, that maximizes the reward. In deep reinforcement learning, the policy is usually parametrized by the weights of a neural net-work that can be optimized with a learning algorithm (e.g. backpropagation, evolutionary methods, etc.). The policy can also be stochastic, that way it's output is a probability distribution from what the agent samples.

#### 1.2.2.2 Trajectory

A *trajectory* is an alternating sequence of states and actions. The first element of a trajectory is the starting state, followed by all the executed actions and their resulting transitions. In light of this, the trajectory of one execution of the environment computes as:

$$\tau = (s_0, a_0, s_1, a_1, ...).$$

#### 1.2.2.3 Return

The *return* is the summation of rewards corresponding to a trajectory. The return can either be finite-horizon undiscounted, which is the cumulative reward

$$R(\tau) = \sum_{t=0}^{T} r_t$$

or infinite-horizon discounted, which is also a summation of rewards, but discounted w.r.t the time step when the action will be taken as

$$R(\tau) = \sum_{t=0}^{T} \gamma^t r_t,$$

where $\gamma \in (0, 1)$ denote the discount factor. This addition is mathematically convenient as we avoid the situation, where the infinite-horizon return never converges to a finite value and it is also intuitively appealing as we include the risk of future actions into the equation.

#### 1.2.2.4 On-Policy Value Function

Assuming that the agent acts according to a policy $\pi$, the *On-Policy Value Function* associates the expected return from a given state to every state in the state space.

$$V^\pi(s) = \underset{\tau \sim \pi}{E}[R(\tau)|s_0 = s]$$

In case the actions are always selected according to the optimal policy, the *Optimal Value Function* computes as:

$$V^*(s) = \max_\pi \underset{\tau \sim \pi}{E}[R(\tau)|s_0 = s]$$

#### 1.2.2.5 On-Policy Action-Value Function

The *On-Policy Action-Value Function* gives the expected return in case an arbitrary action $a$ is selected in state $s$, assuming that the agent will follow the policy $\pi$ after action $a$.

$$Q^\pi(s,a) = \underset{\tau \sim \pi}{E}[R(\tau)|s_0 = s, a_0 = a]$$

In case every action following action $a$ is selected according to the optimal policy, the *Optimal Action-Value Function* is:

$$Q^*(s,a) = \max_\pi \underset{\tau \sim \pi}{E}[R(\tau)|s_0 = s, a_0 = a].$$

#### 1.2.2.6 Advantage

It is not always necessary to precisely determine the absolute goodness of an action, because computing the relative goodness compared to other actions can be sufficient. For a given policy, the *Advantage* function describes the relative goodness of choosing an action in a given state, assuming that after the action is performed, the agent behaves according to the initial policy. It is defined mathematically as:

$$A^\pi(s,a) = Q^\pi(s,a) - V^\pi(s).$$

### 1.2.3 Bellman Equations

The *Bellman Equations* are central to the theory of reinforcement learning. With them we can express the value of a state with the value of other states, so if the value of $s_{t+1}$ is given, we can express the value of $s_t$. The Bellman equation for the value function is given as:

$$V^\pi(s) = \underset{\substack{a \sim \pi \\ s' \sim P}}{E}[r(s,a) + \gamma V^\pi(s')].$$

And for the optimal value function:

$$V^*(s) = \max_a \underset{s' \sim P}{E}[r(s,a) + \gamma V^*(s')]$$

Apparently, the main difference between the value function and the optimal value function is the presence of the maximum over actions, which denotes that an agent must select its' actions such that they maximize the value function.

### 1.2.4 Goal of RL

The core problem in Reinforcement Learning that we intend to solve is the maximization of the expected return of an agent that acts according to a given policy. Let $J(\pi)$ denote the reward function, which is the acquirable return for a given policy $\tau$ as:

$$J(\tau) =: \underset{\tau \sim \pi}{E}[R(\tau)]$$

where $\tau \sim \pi$ expresses the trajectories, which can be traversed by selecting actions according to policy $\pi$. Since the return of these trajectories are given by the environment, the only way to improve the reward function is by modifying the policy in a way where the frequency of high-return trajectories is increased. In other words, high return trajectories should be *reinforced*, hence the name.

Given a stochastic policy in a stochastic environment, the probability of a certain trajectory can be expressed as:

$$P(\tau|\pi) = p_0(s_0) \prod_{t=0}^{T-1} P(s_{t+1}|s_t, a_t)\pi(a_t|s_t)$$

where T is the last time step , the first term of the product is the transition probability of the environment and the second term of the product is the value of the policy function for a given state. If we integrate over all possible trajectories which can be produced by a given policy, we could rewrite the reward function as:

$$J(\tau) =: \underset{\tau \sim \pi}{E}[R(\tau)] = \int_\tau P(\tau|\pi)R(\tau).$$

This way reinforcement learning can be formulated as an optimization problem:

$$\pi^* = \arg \max_\pi J(\pi).$$

### 1.2.5 Policy Gradient Methods

During a *Policy Gradient Method* [33], the stochastic policy is optimized in a direct manner by using a parameterized function approximator. This could be for example a neural network, which in such case is called a policy network. By feeding the observed state of the environment into the network characterized by policy parameters, the output can be computed as a probability distribution over possible actions. Consequently, the policy parameters can be updated approximately proportionally to the gradient:

$$\boldsymbol{\theta_{t+1}} = \boldsymbol{\theta_t} + \alpha \widehat{\nabla J(\boldsymbol{\theta_t})},$$

where $\boldsymbol{\theta_t}$ denote the policy parameters at time step $t$, $\alpha$ is the step size and the last term is the stochastic estimate, whose expectation approximates the gradient of the performance measurement w.r.t. the policy parameters. Algorithms that utilize this gradient ascent method for efficiently optimizing the policy belong to the family of *Policy Gradient Methods*.

#### 1.2.5.1 Policy Gradient Theorem

While a policy function is being trained, it's performance not only rely on the selection of actions, but also the distribution of states where the policy selects those actions. While for a given state, the influence of policy parameters on a chosen action can easily be computed (and therefore the observed reward), the influence on state distribution changes as a function of the environment, so typically in an unknown manner. It is challenging to estimate the performance of the policy w.r.t the policy parameters as distribution shifts in the state space may occur.

Fortunately, the *Policy Gradient Theorem* provides a theoretical answer to this problem with an analytical expression concerning the gradient of the policy performance by rewriting the derivative of the objective function to not involve the derivative of the state distribution, hence simplifying the computation of the gradient:

$$\nabla J(\boldsymbol{\theta}) \propto \sum_s \mu(s) \sum_a q_\pi(s,a) \nabla_{\boldsymbol{\theta}} \pi(a|s,\boldsymbol{\theta}) \tag{1.1}$$

where $\mu$ denote the state distribution.

### 1.2.5.2 REINFORCE: Monte Carlo Policy Gradient

One of the core policy gradient algorithms is *REINFORCE* [38], which provides a way to obtain samples such that the expectation of the sample gradient is proportional to the actual gradient of the performance function w.r.t. the parameters of the function approximator neural network. The sample gradients are sufficient to be proportional to the real gradients because the arbitrarily chosen $\alpha$ step size incorporates the rest by being able to absorb any constant of proportionality. *REINFORCE* supports the policy gradient theorem with a way of sampling that approximates the expectation of the right hand side (1.1), which is the sum over all states weighed by how often that state occurs given the target policy. Following this target policy, the agents visit these states with this proportion, thus we can rewrite the equation as:

$$\nabla J(\boldsymbol{\theta}) \propto \sum_s \mu(s) \sum_a q_\pi(s,a) \nabla_{\boldsymbol{\theta}} \pi(a|s,\boldsymbol{\theta})$$
$$= E_\pi \left[ \sum_a q_\pi(S_t, a) \nabla_{\boldsymbol{\theta}} \pi(a|S_t, \boldsymbol{\theta}) \right]$$

where $S_t$ mark the sampled actions. Now we want to replace the sum over action with the sampled action $A_t$. In order to accomplish this, we need to introduce the probability according to which we select each action given the policy. We are going to arrange this by multiplying and then dividing with this probability:

$$\nabla J(\boldsymbol{\theta}) = E_\pi \left[ \sum_a \pi(a|S_t, \boldsymbol{\theta}) q_\pi(S_t, a) \frac{\nabla_{\boldsymbol{\theta}} \pi(a|S_t, \boldsymbol{\theta})}{\pi(a|S_t, \boldsymbol{\theta})} \right]$$
$$= E_\pi \left[ q_\pi(S_t, A_t) \frac{\nabla_{\boldsymbol{\theta}} \pi(A_t|S_t, \boldsymbol{\theta})}{\pi(A_t|S_t, \boldsymbol{\theta})} \right].$$

This way we replaced $a$ by the sample $A_t \sim \pi$. In the next step we will use that $E_\pi[R(\tau_t)|S_t, A_t] = q_\pi(S_t, A_t)$:

$$\nabla J(\boldsymbol{\theta}) = E_\pi \left[ R(\tau_t) \frac{\nabla_{\boldsymbol{\theta}} \pi(A_t|S_t, \boldsymbol{\theta})}{\pi(A_t|S_t, \boldsymbol{\theta})} \right]$$

The fractional term is often referred to as the *eligibility vector*, where the numerator expresses the gradient of the probability of selecting an action and the denominator denotes the probability of selecting the action. This vector can be written in a more compact form of $\nabla_{\boldsymbol{\theta}} \ln \pi(A_t|S_t, \boldsymbol{\theta_t})$ using the identity $\nabla \ln x = \frac{\nabla x}{x}$. We can sample this vector at each time step during interaction with the environment, giving an estimate to the performance measure of the current policy. It is worth noting, that *REINFORCE* is a Monte Carlo method, because to know the value of $R(\tau_t) = G(t)$, a complete trajectory needs to be sampled using a single policy. Then, the product of the two terms is used to update the policy parameters after each episode with the formula of stochastic gradient ascent:

$$\boldsymbol{\theta_{t+1}} = \boldsymbol{\theta_t} + \alpha R(\tau_t) \frac{\nabla_{\boldsymbol{\theta}} \pi(A_t|S_t, \boldsymbol{\theta_t})}{\pi(A_t|S_t, \boldsymbol{\theta_t})}$$

Intuitively, these updates increase the chance of taking an action when the observed return from a given state is high. The updates are proportional to the yielded return and inversely proportional to the action probability. The latter is important, because this way the agent does not get biased towards actions that are chosen more frequently. Below you may find the pseudocode of *REINFORCE*.

---

**Algorithm 1** REINFORCE: Monte Carlo Policy Gradient
---
**Input:** a differentiable policy parameterization $\pi(a|s, \boldsymbol{\theta})$
**Initialize** the policy parameters $\boldsymbol{\theta} \in R^{d'}$
**while** *True* **do**
    **foreach** $(S_0, A_0, R_1, ..., S_{T-1}, A_{T-1}, R_T) \sim \pi_{\boldsymbol{\theta}}$ **do**
        **for** *t=0,...,T-1* **do**
            $R \leftarrow$ *return from step t*
            $\boldsymbol{\theta} \leftarrow \boldsymbol{\theta} + \alpha \gamma^t R \nabla_{\boldsymbol{\theta}} \ln \pi(A_t|S_t, \boldsymbol{\theta})$
        **end**
    **end**
**end**

---

An additional extension to *REINFORCE* and other policy gradient algorithms is to include a comparison of the value function with a baseline function, which can be any arbitrary function as long as it does not depend on $a$. This way the Policy Gradient Theorem can be rewritten in the form of:

$$\nabla J(\boldsymbol{\theta}) \propto \sum_s \mu(s) \sum_a \Big( q_\pi(s,a) - b(s) \Big) \nabla_{\boldsymbol{\theta}} \pi(a|s, \boldsymbol{\theta}),$$

where $b(s)$ is the baseline function. The equations remains valid due to the fact that the value of the subtracted quantity is still zero.

$$\sum_a b(s) \nabla_{\boldsymbol{\theta}} \pi(a|s, \boldsymbol{\theta}) = b(s) \nabla_{\boldsymbol{\theta}} \sum_a \pi(a|s, \boldsymbol{\theta}) = b(s) \nabla_{\boldsymbol{\theta}} 1 = 0$$

Therefore it is valid to modify the parameters as:

$$\boldsymbol{\theta_{t+1}} = \boldsymbol{\theta_t} + \alpha \Big( R(\tau_t) - b(S_t) \Big) \frac{\nabla_{\boldsymbol{\theta}} \pi(A_t|S_t, \boldsymbol{\theta_t})}{\pi(A_t|S_t, \boldsymbol{\theta_t})}.$$

Acknowledging that the baseline does not affect the expected value of the policy gradient update, this algorithm remains a strict generalization of *REINFORCE*. The motivation behind the introduction of such a baseline function is that with a properly selected $b(s)$, the variance of updates can be reduced, which results in a better convergence overall. A good choice for the baseline function is the value function, because this way we are given the advantage function, which expresses the relative return for a particular action compared to what is usually observed from the current state.

$$A^{\pi_{\boldsymbol{\theta}}}(s,a) = Q^{\pi_{\boldsymbol{\theta}}}(s,a) - V^{\pi_{\boldsymbol{\theta}}}(s)$$

Since $V(s)$ is the expected return from state $s$, having it subtracted from $Q(s,a)$, we remove the dependency from all future states along with their rewards.

### 1.2.5.3 Proximal Policy Optimization

Vanilla policy gradient algorithms utilize gradient ascent to take steps in the steepest direction on the loss surface using a first order optimizer. However, first order methods can sometimes be inaccurate near curved areas, which in case of overconfidence may lead to performance collapse. The main reason behind this phenomenon is that a bad step in the policy update not only lower the performance, but it also has a serious effect on the state distribution, thus it is essential to keep the learning rate coefficient low in order to ensure stable policy updates.

The first proposal for solving this problem is Trust Region Policy Optimization (TRPO) [28], where a constraint is introduced on the space of policy updates. When searching for the next, updated policy in this region, the constraint ensures that the new policy will not be drastically far from the current policy's behaviour. This is achieved by taking the Kullback-Leibler divergence of the parameters of the current and new policies.

In practice, unfortunately TRPO is a computationally expensive algorithm to perform. To address this, *Proximal Policy Optimalization* (PPO) methods [30] were proposed, which contrary to TRPO, are first order methods. The simplest form of PPO introduces a clip term in the objective function (also called surrogate loss), to remove incentives for a too big policy update:

$$J(s,a,\boldsymbol{\theta_k},\boldsymbol{\theta}) = min\left(\frac{\pi_{\boldsymbol{\theta}}(a|s)}{\pi_{\boldsymbol{\theta_k}}(a|s)}A^{\pi_{\boldsymbol{\theta_k}}}(s,a), \text{clip}\left(\frac{\pi_{\boldsymbol{\theta}}(a|s)}{\pi_{\boldsymbol{\theta_k}}(a|s)}, 1-\epsilon, 1+\epsilon\right)A^{\pi_{\boldsymbol{\theta_k}}}(s,a)\right) \quad (1.2)$$

where $\boldsymbol{\theta_k}$ is the policy at time step $k$, according to which we arrived to the current state and $\boldsymbol{\theta}$ is the new policy. From the second term in the equation, we can see that if the difference between the previous and the new policy is greater than a parameter $\epsilon$, the new policy will not be further rewarded. Then, we choose the smaller between the clipped and the unclipped objective, so they form a lower bound on the unclipped objective. It is worth noting, that the probability ratio is clipped, whether the advantage is positive or negative.

During PPO, we calculate the surrogate loss after having collected a given number of trajectories and then we use an optimizer such as stochastic gradient descent to iteratively update the policy. The pseudocode of the algorithm looks like the following:

---
**Algorithm 2** Proximal Policy Optimization (Actor-Critic Style)
---
**for** *iteration=1,2,...* **do**
    **for** *actor=1,2,...,N* **do**
        Run policy $\pi_{\boldsymbol{\theta}_{\text{old}}}$ in environment for T time steps
        Compute advantage estimates $\hat{A}_1, ..., \hat{A}_T$
    **end**
    Optimize the surrogate J w.r.t. $\boldsymbol{\theta}$, with K epochs and minibatch size M $\leq NT$
    $\boldsymbol{\theta}_{\text{old}} \leftarrow \boldsymbol{\theta}$
**end**

---

## 1.3 Neural Architecture Search

In the past decade, deep neural networks had been significantly improved in terms of performance. This was primarily due to all the theoretical breakthroughs of recent years and the continuously increasing computing power available at hand. When putting these novel algorithms and techniques into practice, people often rely on heuristics proposed by domain experts. Only a very small percentage of society have the necessary experience and knowledge to utilize deep learning methods to their full potential, so the majority of people build their models based on biases. In the context of image recognition for example, it is a standard approach to load a predesigned architecture such as ResNet [11] and try to tweak the parameters until sufficient result is achieved. With the help of the increasing computing power, the field of Automated Machine Learning (AutoML) addresses this issue, by providing different optimizations and parameter searching methods for automating the entire machine learning pipeline. *Neural architecture search (NAS)* is a subfield of AutoML focused on finding the optimal build blocks of neural networks and their topology for a specific task.

Neural architecture search is an actively researched area and a wide variety of efficient methods had been proposed in recent years. One major category of algorithms include a controller model, that designs and evaluates architectures with different sets of hyperparameters and use their validation accuracies to improve the controller's performance of designing networks automatically. In this case, many approaches exist for the learning of the controller, e.g. it could be an evolutionary algorithm [31] or in case of our work, reinforcement learning [39, 25]. Another important approach is the application of continuously differentiable methods, such as Differentiable Architecture Search (DARTS) [19], where the structure of the network is continuously being relaxed while backpropagating the error to it.

In the following two chapters, we will briefly discuss and highlight the characteristics of existing NAS methods using reinforcement learning.

### 1.3.1 NAS with Reinforcement Learning

Barret Zoph and Quoc Le were the first to publish a reinforcement learning based neural architecture search method [39] that not only had a comparable performance to neural networks designed by human experts, but the discovered convolutional network actually achieved state-of-the-art accuracy on CIFAR-10 [15] and also converged faster than it's contestants. They also evaluated the algorithm on recurrent cell structures similar to an LSTM's.

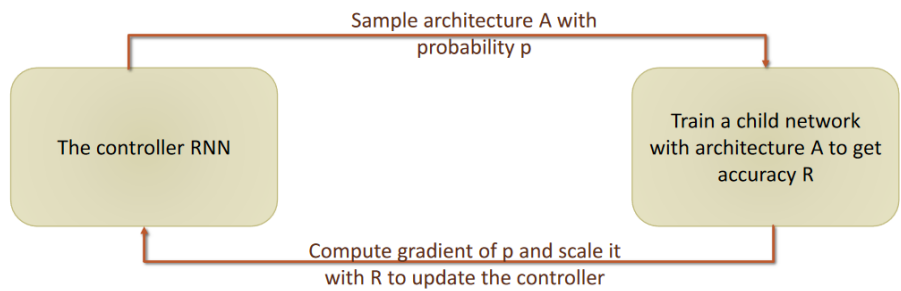

**Figure 1.8:** Overview of neural architecture search with reinforcement learning [2]

The controller - built up by LSTM cells - generates the architectures (child models) that are individually trained for a given number of epochs. After the training is done, the validation accuracies of the child models serve as a reward signal for the controller, which utilizes the REINFORCE rule to update its' parameters.

#### 1.3.1.1 Generation of architectures

Their proposed controller RNN had two layers of cells, each of them having 35 LSTM units. This model generated the hyperparameters defining the child architectures layer by layer in a given sequence by generating a token for every time step. The controller continued this process iteratively, until the children consisted of a predefined number of layers, then the neural networks were constructed and trained.

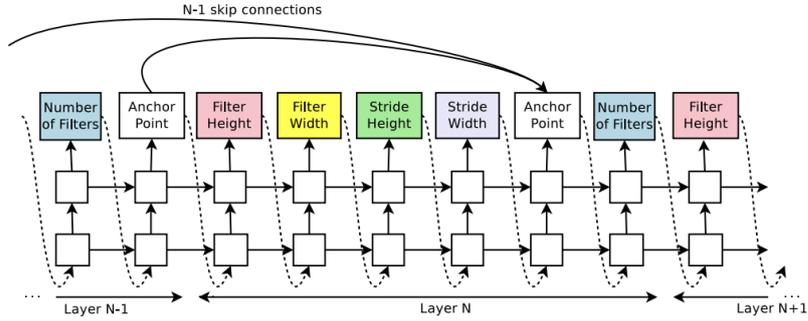

**Figure 1.9:** The process of sequential feature generation and prediction of skip connections [39]

To increase the search space of architectures, more types of cells and skip connections were added to the building blocks, patterned after GoogleNet [34] and ResNet [11].To be able to efficiently predict where to use skip connections, the controller uses the attention mechanism [5] to select the endpoints of the connections (as shown in figure 1.9). For the $N$th layer, it takes into consideration the sigmoid of $N-1$ anchor points in order to choose the previous layers that need to be connected to the current one. Every sigmoid is dependent on the current and the previous hidden state of the controller.

$$P(\text{Layer j is an input to layer i}) = \text{sigmoid}(\boldsymbol{v^T}\tanh(\tanh(\boldsymbol{W_{prev}} * h_j + \boldsymbol{W_{curr}} * h_i)),$$

where $\boldsymbol{v^T}, \boldsymbol{W}_{\text{prev}}$ and $\boldsymbol{W}_{\text{curr}}$ are the trainable parameters with $h_j$ representing the hidden state of the controller at an achor point for the $j$th layer, which ranges from 0 to $N-1$. We then sample from this multivariate distribution of sigmoids to decide which previous layers are going to be used to be the input of the current layer. The inputs of every layer are concatenated in the depth dimension, which results in further complications, for which the following additions are noteworthy:

- If a layer is not connected to anything, it is connected directly to the input layer.

- When reaching the final layer during the construction of the network, every layer that had no output, are concatenated and given to the input of the last hidden state.

- If certain layers vary in spatial dimensions, zero padding is applied to make them match.

### 1.3.1.2 Training the controller with REINFORCE

In order to formulate the controller as a reinforcement learning agent, the tokens produced by the model are corresponded to actions $(a_1, a_2, ..., a_t)$. A generated child architecture is represented as a trajectory, which is a series of actions proportional to the number of layers. Having trained the generated neural network, a reward is assigned to the trajectory by taking the validation accuracy of the child. This accuracy will be the driving force of the reinforcement learning algorithm, by using it as a reward signal to train the $\boldsymbol{\theta_c}$ parameters of the controller. Therefore the objective of the controller is to maximize the following reward:

$$J(\boldsymbol{\theta_c}) = E_{P(a_{1:T};\boldsymbol{\theta_c})}[R]$$

The reward signal $R$ is not differentiable, but it is possible to utilize policy gradient algorithms to optimize $\boldsymbol{\theta_c}$. For this they applied the previously introduced REINFORCE algorithm:

$$\nabla_{\boldsymbol{\theta_c}} J(\boldsymbol{\theta_c}) = \sum_{t=1}^{T} E_{P(a_{1:T};\boldsymbol{\theta_c})} \left[ \nabla_{\boldsymbol{\theta_c}} \log P(a_t|a_{(t-1):1};\boldsymbol{\theta_c}) R \right],$$

which can be approximated empirically by:

$$\frac{1}{m} \sum_{k=1}^{m} \sum_{t=1}^{T} \nabla_{\boldsymbol{\theta_c}} \log P(a_t|a_{(t-1):1};\boldsymbol{\theta_c}) R_k,$$

where $m$ denotes the generated architectures in one batch and $T$ marks the required number of hyperparameters of the architecture predicted by the controller. The above approximation is an unbiased, but high-variance estimate of the real gradient, hence a baseline function is introduced to reduce variance as:

$$\nabla_{\boldsymbol{\theta_c}} J(\boldsymbol{\theta_c}) = \frac{1}{m} \sum_{k=1}^{m} \sum_{t=1}^{T} \nabla_{\boldsymbol{\theta_c}} \log P(a_t|a_{(t-1):1};\boldsymbol{\theta_c}) (R_k - b),$$

where $b$ is the baseline function. In the paper they proposed an exponential moving average as a baseline, which is not dependent on the current action, thus bias is still not included in the gradient estimate.

Obtaining a sufficient number of rewards by interacting with the environment is a very expensive task in terms of computational power, as every child model has to be trained until convergence. The process is sped up by significant parallelization and asynchronous updates with the introduction of a parameter server on which multiple controller replicas are run. After these replicas have generated and trained the child networks in parallel, they send the calculated gradients to the parameter server, in order to update the parameters of each controller replica.

### 1.3.1.3 Search space for image recognition

In their setup, the search space for image recognition consists of convolutional layers, ReLU nonlinearities, batch normalizations, skip connections between layers and finally fully connected layers. For every layer, four hyperparameters are chosen by the controller with the following possible values:

- filter height: $[1, 3, 5, 7]$
- filter width: $[1, 3, 5, 7]$
- number of filters: $[24, 36, 48, 64]$
- stride: $[1, 2, 3]$

That means with L number of layers the number of different configurations was $(4*4*4*3)^L * 2^{(L(L-1)/2)}$.

### 1.3.2 Efficient Neural Architecture Search via Parameter Sharing

The main contribution of Efficient Neural Architecture Search (ENAS) [25] is overcoming the biggest weakness of the standard reinforcement learning-based neural architecture search, by decreasing the time and resources needed for training. Their best architecture was on par with the original NAS performance, while they only used a single GPU and the training only took days instead of weeks. This is achieved by significantly decreasing the training times of child models with the introduction of a parameter sharing method among sampled architectures. As the algorithm reuses the weights of previously trained models - after a certain number of iterations - even the untrained networks perform relatively well. Although different models may utilize their weights differently due to architectural differences, their proposed algorithm is not only capable of searching for complex architectures, but also delivers strong empirical results. In terms of numbers, their architectures show comparable performance to the original RL-based method, with their recurrent model even achieving higher accuracy on the Penn Treebank dataset [21]. Their significant contribution however, is that they achieved a three orders of magnitude speed-up in terms of training times as one day of training on one GPU was sufficient for training the searching algorithm.

#### 1.3.2.1 Generation of architectures

The main idea of ENAS is that the search space of the problem can be represented as a single directed acyclical graph (DAG), where the nodes mark the specific operations and the edges represent the flow of information. Therefore, the problem of finding well-performing architectures can be formulated as sampling subgraphs from this large computational graph (as shown in figure 1.10).

The controller samples from the computational graph similarly to how the original RL-based method works. It's initial input is an empty embedding, which later uses the previous outputs as input for the current iteration in an autoregressive manner. Every output represent a decision in the sampled subgraph, which on one hand is selecting a node where computations are performed and on the other hand selecting the edges from where the nodes receive information. From all the possible values, these are selected by a softmax classifier.

#### 1.3.2.2 Search Space

ENAS was evaluated both for sequential modelling and image recognition on Penn Treebank and CIFAR-10 respectively. In case of recurrent networks, they were able to search for both the cell topology and all the executed operations within. During sampling, the controller selects an activation function (tanh, ReLU, identitiy, sigmoid) for each node in a cell and then selects an input for the node, which could alternatively be the input or the previous state of the cell. The cell's output then computes as the average of all the nodes that were not selected as an input to nodes. For image recognition, ENAS is capable of searching both for convolutional modules and convolutional architectures. The first is called micro search, whereas the latter is called macro search.

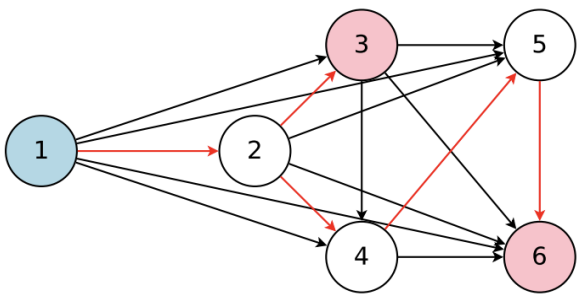

**Figure 1.10:** The directed acyclical graph representing the entire search space. In this example, the red arrows denote a sampled model, the blue node is the input, while the red nodes are the outputs of the network. [25]

In micro search, rather then designing the whole network, they search the internal structure of a general submodule and then they connect more of them together to form a network.

During the search for convolutional architectures in macro search - similarly to recurrent cell search - the controller predicts the operation for a given node and all the previous nodes from where it's inputs are derived from. Since it is allowed to sample inputs from multiple nodes to one node, the inputs are concatenated depthwise [8] before they are fed to the node. The operations performed in a node are chosen from 6 types of blocks, which included pooling, separable convolution and other types of operations.

### 1.3.2.3 Training

During training, two sets of hyperparameters need to be optimized. The weights of the controller LSTM denoted by $\boldsymbol{\theta}$ and the shared parameters of the child models denoted by $\boldsymbol{\omega}$. A complete training of the ENAS algorithm consists of an alternating sequence of training these two sets of parameters. In the first phase, a sampled child model's parameters are trained with backpropagation, then in the second phase, the acquired validation accuracy is used to update the controller parameters.

When training the shared parameters, at first $M$ child architectures are sampled from the controller's fixed policy $\pi(m|\boldsymbol{\theta})$ and have stochastic gradient descent performed on its' parameters $\boldsymbol{\theta}$ in order to minimize the expected loss function. This loss function $L(m;\boldsymbol{\omega})$ is a negative log likelihood function, which is a form of cross-entropy loss computed on every batch of the training data. This way the gradient computes as:

$$\nabla_{\boldsymbol{\omega}} E_{\boldsymbol{m} \sim \pi(\boldsymbol{m};\boldsymbol{\theta})}[L(\boldsymbol{m};\boldsymbol{\omega})] \approx \frac{1}{M} \sum_{i=1}^{M} \nabla_{\boldsymbol{\omega}} L(\boldsymbol{m}_i, \boldsymbol{\omega}),$$

where $\boldsymbol{m}_i$ is a sampled child architecture. This gradient estimate is unbiased, however, it has a higher variance estimate than the standard SGD gradient in which case $\boldsymbol{m}$ is fixed. Surprisingly, Pham et al. discovered, that $M = 1$ works sufficiently, so in other words, training the shared parameters $\boldsymbol{\omega}$ can be done by sampling any single model from $\pi(m|\boldsymbol{\theta})$ and hence computing the gradient.

When training the controller, the shared parameters of the child models $\boldsymbol{\omega}$ are fixed and the controller parameters $\boldsymbol{\theta}$ are updated to maximize the expected reward:

$$\nabla_{\boldsymbol{\omega}} E_{\boldsymbol{m} \sim \pi(\boldsymbol{m};\boldsymbol{\theta})}[R(\boldsymbol{m};\boldsymbol{\omega})],$$

where $R(\boldsymbol{m};\boldsymbol{\omega})$ is the accuracy of a child. To prevent the controller from overfitting on the training data and induce generalization, this accuracy is always computed on the validation data set. Updating the parameters with this reward signal is done exactly like in the case of the original NAS with reinforcement learning, using the REINFORCE algorithm.

# Chapter 2

# Methods

## 2.1 Proposed work

REINFORCE, the reinforcement learning algorithm driving ENAS is a fairly simply policy gradient method. In this work, we would like to improve ENAS in terms of sample-efficiency, by modifying the learning of the controller to Proximal Policy Optimization. The main goal of this paper is to show that a controller trained with PPO is able to achieve on par performance with REINFORCE, while seeing fewer samples. In order to validate our hypothesis, we created our own implementation of ENAS, primarily based on the original publication and the corresponding implementation [1]. Since the learning of the controller is not closely related to the type of searched architectures, we decided to narrow our scope for image classification, using a macro search space, because if in this case the learning of the controller improved, it would have a similar impact on learning in different search spaces.

## 2.2 Our implementation of ENAS

### 2.2.1 Child models

The fundamental building units of the child models are the blocks. During our experiments, we used 6 types of blocks from which the controller could sample layer by layer. Apart from this, similarly to the ENAS implementation, we placed reduction blocks with a stride of 2 to certain locations in the network, to reduce the spatial dimensions of the channels. If a skip connection is between layers with different spatial dimensions, we apply downsampling for them to match. Also, after every reduction layer, we doubled the number of filters in the following layers. It is worth noting, that the number of reduction blocks was independent of the number of searched layers as it was always 2, and their location was at the 1/3 and 2/3 of the network. The internal operations of the blocks we used could be chosen arbitrarily, however we induced a bias in the process by using well-working convolutional building blocks, hence narrowing down the search space. The convolutional blocks in our model all follow the ReLU-convolution-batchnorm structure [14] and consist of the following blocks as shown in Figure 2.1:

- convolution with kernel size 3x3 (2.1 top-right)

- convolution with kernel size 5x5 (2.1 bottom-right)

- depthwise separable convolution 3x3 (2.1 top-middle)
- depthwise separable convolution 5x5 (2.1 top-left)
- average pooling 3x3 (2.1 bottom-middle)
- max pooling 3x3 (2.1 bottom-left)

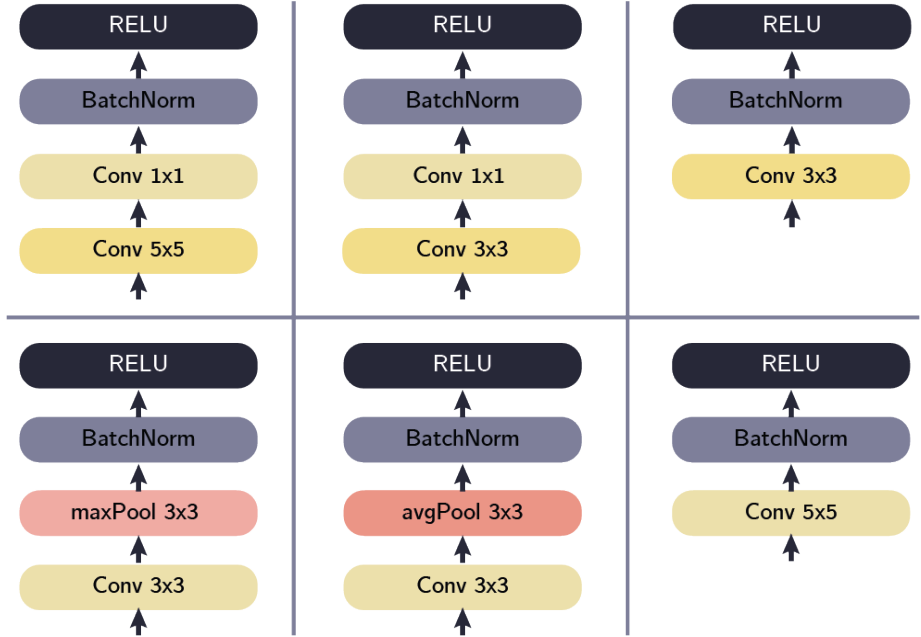

**Figure 2.1:** The internal structure of the 6 blocks from which the controller samples

In light of this, the size of the macro search space for convolutional architectures computes as $6^L * 2^{\frac{L(L-1)}{2}}$ for $L$ layers. As an example, if $L = 10$, the number of possible architectures is $2.1 * 10^{21}$.

At the beginning of the network, before the predicted layers, a convolutional layer receives the input image. After the predicted layers, there is a global average pooling layer which calculates the average output of each feature map in the previous layer and therefore helps to reduce the number of parameters in the dense fully connected layer, where the number of neurons was equal to the number of classes in the dataset.

We performed the training of the child models using stochastic gradient descent with a cosine annealing scheduled learning rate [20] which in many cases, provided better empirical results.

### 2.2.2 Controller

The process of sampling a layer for a child model (as shown in Figure 2.2) looks like the following:

1. Forward propagate the input into the LSTM and sample from the output logit distribution. This will be the selected block type.

2. Pass the selected block to an embedding and forward propagate again through the LSTM.

3. The output received in 2.2. is added to the array of anchor points, to which we will have the possibility to connect skip connections later on.

4. Use an attention mechanism to predict skip connections to previous anchor points from the current layer. If it is the first layer, we skip this step.

5. The input of the next layer will be the mean of anchor points corresponding to the predicted skip connections.

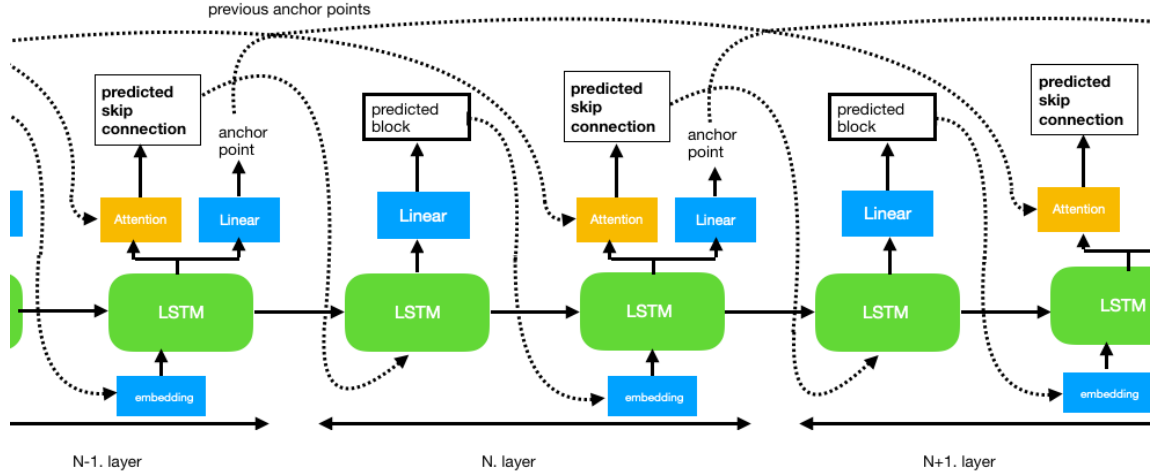

**Figure 2.2:** The process of sampling layers and skip connections

### 2.2.3 Training

The learning of the controller in case of REINFORCE is conducted similarly to the original version of ENAS.

1. Sample $M$ architectures in every epoch with a fixed policy.

2. For every architecture, perform a whole pass through on the training dataset and take their average performance on the validation dataset.

3. Substract the average of the validation accuracies of the last 5 controller epochs from the previously acquired value to see how much the policy has improved in this step. We call the hence acquired value reward. It is important to note, that it does not completely reflect the improvement, because the shared parameters are being trained as well.

4. Multiply the reward with the log probabilities of the controller's decisions and we get loss function of REINFORCE

    4.1. By adding the entropy of the controller's decisions to the loss, exploration is encouraged.

5. Update the parameters of the controller using gradient ascent on the loss function.

To summarize the REINFORCE-based neural architecture search, we present the pseudocode as well:

**Algorithm 3** Neural architecture search using REINFORCE
---
controller = $C$
**foreach** *Controller epoch* **do**
    SAMPLE $M$ child architecture from $C_{old}$
    **foreach** *child $m_i$ in $M$* **do**
        TRAIN $m_i$ for *child_epochs* number of epochs
        $reward$ = calculate the validation accuracy
        $baseline$ = calculate previous controller epoch mean validation accuracy
        $loss = (sampled\_logprobs * (reward - baseline)) + sampled\_entropies$
        Perform gradient ascent on loss function to update $C$
    **end**
**end**
---

During an architecture search with this method, the majority of time and computational resources are spent with collecting the reward signal by evaluating child models. Previous research in this area was primarily focused on making this evaluation faster, however we decided to try to improve the process by making the controller more sample-efficient. The proposed way of doing this is by modifying the learning algorithm of the controller to Proximal Policy Optimization.

## 2.3 Motivation behind PPO

Policy gradient methods have achieved several recent breakthroughs in deep reinforcement learning [29]. However, it should not be ignored, that training agents via policy gradient is often a challenging task, because they are very sensitive to hyperparameters. One of these critical parameters is the step size. If it is too low, the learning of the agent is extremely slow, while if it is too large, the model often diverges due to overconfident policy updates, thus slowing or completely preventing learning. Policy gradient methods often have a very poor sample efficiency compared to other methods like Q-learning with different kinds of experience replay [27, 3]. Recent policy gradient approaches like TRPO and ACER [36] partly eliminated these problems by introducing constraints into the policy updates, but unfortunately it does come with a cost. These methods cannot be used generally in many use-cases and they are hard to implement. Contrary to this, Proximal Policy Optimization is relatively easy to implement and tune. It also provides a strong empirical performance and sample complexity. It's novel loss function allows relatively big parameter updates on every sample, while ensuring that the deviation from the previous policy is relatively small with it's loss clipping, thus preventing the agent from a sudden policy collapse.

## 2.4 Modifying the controller to PPO

The main difference between a vanilla policy gradient like REINFORCE and PPO is the way we compute the loss and update the policy's parameters. In the implementation of Proximal Policy Optimization in every controller epoch, we store all actions and the corresponding information, which was related to the controller's decision. Normally this would be the state, the action, the next state and the reward but in our case, we also have to store the LSTM's hidden states and the previous anchor point in the current child. We need these information to compare the log probabilities of both the policy from which we previously sampled the architectures(old controller)and the policy that we are updating(new controller).

For calculating the Advantage, most PPO algorithms utilize an actor-critic model, where the value function is also approximated with a neural network. Our case is somewhat special, because the reward is associated with the entire trajectory corresponding to the generation of that child model, thus we use the validation accuracy as value and then subtract the baseline (moving average validation accuracy of previous controller epochs) given the advantage from the current validation accuracy.

We iterate through every action selected in the last controller epoch $K$ times, compute the surrogate loss function of Proximal Policy Optimization (Equation 1.2) and update the new policy's parameters with gradient ascent on this loss. Having updated the policy, we simply copy the updated policy's weights into the old policy and repeat the process.

The pseudocode of the proposed neural architecture search with PPO looks like the following:

---
**Algorithm 4** Neural architecture search using PPO
---
Old controller = $C_{old}$
New controller = $C_{new}$
COPY parameters of $C_{old}$ to $C_{new}$
Memory = *mem*
**foreach** *Controller epoch* **do**
    SAMPLE $M$ child architecture from $C_{old}$
    **while** *sampling* **do**
        SAVE *input*, *output*, *hidden_states*, *anchor_points* to *mem*
    **end**
    **foreach** *child $m_i$ in $M$* **do**
        TRAIN $m_i$ for *child_epochs* number of epochs
        Calculate *reward*
        SAVE *reward* to *mem*
        **foreach** *K_epoch* **do**
            Calculate *log_probs* from $C_{old}$ with information from *mem*
            Calculate *log_probs* from $C_{new}$ with information from *mem*
            Calculate *surrogate_loss*
            Perform gradient ascent on loss function to update $C_{new}$ parameters
        **end**
        COPY $C_{new}$ parameters to $C_{old}$
    **end**
**end**

# Chapter 3

# Experiments

Reproducibility is a recurring issue in machine learning research, especially in the field of neural architecture search. Marius Lindauer and Frank Hutter very recently put forward some best practices [18], which should be kept in mind in order to ensure reproducibility. In this chapter, we present the experiments we conducted and our results, while doing our best to follow the proposed guidelines.

## 3.1 Training details

### 3.1.1 Dataset

We used CIFAR-10 for image classification during our experiments, because it is one of the standard benchmark datasets for neural architecture search. The dataset consists of 60000 32x32 RGB pictures, which can be categorized into 10 categories. We used 50000 images for training and 10000 for the validation of architectures. We performed the same preprocessing steps as in the ENAS and the original NAS paper, which consisted of normalization, random horizontal flipping, padding the images to 40x40 and then randomly cropping them back to their original size.

### 3.1.2 Hyperparameters

At first we tried to use the hyperparameters proposed in ENAS, however due to computational limitations, we decreased the size of the network and reduced the number of filters for each layer, thus speeding up the training process. We also noticed that decreasing the size of the controller and increasing the learning rate improved our algorithms, both in case of REINFORCE and PPO. This difference might derive from the differences in the implementation and the training environment.

| Parameter | Value |
|---|---|
| **Child models** | |
| Number of layers in child models | 10 |
| Number of reduction blocks | 2 |
| SGD: learning rate | 0.05 |
| Initial out filters | 24 |
| SGD: Momentum | 0.5 |

| Continuation of Hyperparameters | |
| --- | --- |
| **Parameter** | **Value** |
| SGD: l2 weight decay | 2E-04 |
| Dropout probability | 0.8 |
| Cosine LR Scheduler: $t_0$ | 10 |
| Cosine LR Scheduler: $eta\_min$ | 0.001 |
| Cosine LR Scheduler: $t\_mult$ | 2 |
| **Controller** | |
| Number of LSTM layers | 2 |
| Number of LSTM units per layer | 20 |
| $Tanh\_const$ | 2.5 |
| Temperature | 5 |
| Initial weights | [-0.1, 0.1] |
| ADAM: learning rate | 0.006 |
| ADAM: betas | (0.0, 0.999) |
| ADAM: epsilon | 1E-03 |
| Entropy weight | 0.01 |
| **Training** | |
| Controller epochs | 150 |
| Child models per controller epoch | 3 |
| epoch per child models | 1 |
| PPO: $eps\_clip$ | [-0.2,0.2] |
| PPO: $K\_epochs$ | 10 |

**Table 3.1:** The hyperparameters used during the experiments

### 3.1.3 Training environment

For initial testing and hyperparameter tuning, we used Google Colaboratory[1] with GPU instances. This was particularly useful during the development phase, because we could train multiple controllers in parallel, using different hyperparameter settings. The final models were trained on the Amazon AWS Sagemaker service[2], on p2.xlarge EC2 instances for 12 hours. These machines are equipped with NVIDIA K80 GPUs, each with 2,496 parallel processing cores and 12GiB of GPU memory.

Although the original implementation was written in Tensorflow[3] and there were not any properly working, open-source PyTorch implementations of ENAS for macro search on image classification, we chose PyTorch for the verification of methods, because it gives a greater flexibility when debugging the models. For logging, storing hyperparameter settings and visualizing experiments, we used Tensorboard[4]. The implementation used for the verification of the proposed work can be found in an open-source repository[5] on GitHub.

---

[1] https://colab.research.google.com/
[2] https://aws.amazon.com/sagemaker/
[3] https://www.tensorflow.org/
[4] https://www.tensorflow.org/tensorboard/
[5] https://github.com/attilanagy234/ENAS-pytorch

### 3.1.4 Evaluation

During evaluation, we compare the proposed PPO architecture search to our REINFORCE search. As a control method, we compare both of the algorithms to an untrained controller, which also samples from the shared parameter space. This random search is necessary to ensure that the performance is not only increasing because of a well-defined search space, but also because we successfully trained the controller.

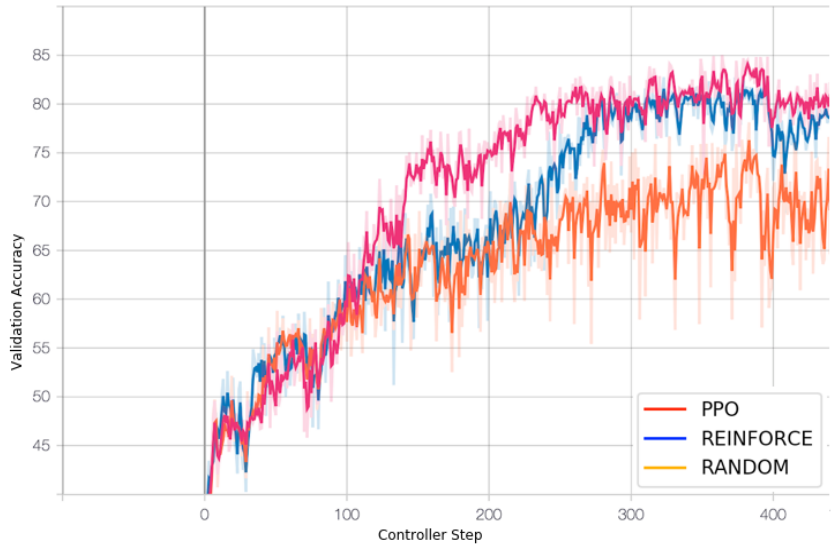

**Figure 3.1:** The validation accuracies over time of the controllers trained with the 3 examined methods.

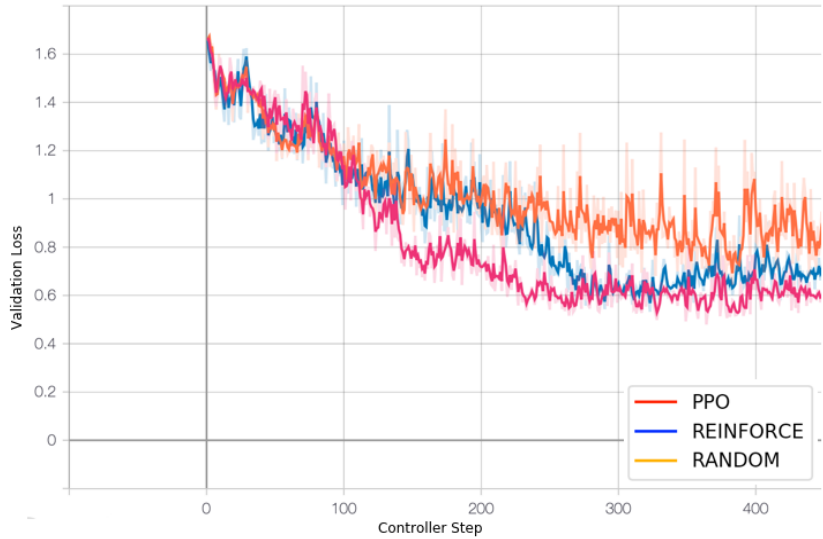

**Figure 3.2:** The loss over time of the controllers trained with the 3 examined methods.

### 3.1.4.1 Comparison of the generated child architectures

Having trained the controllers, we choose the best architectures found during the search and retrain them from scratch. We retrained each of the models on 50000 samples from CIFAR10 for 300 epochs and then tested their accuracy on the remaining set of images. Firstly, we used the same hyperparameters as during the search, then we doubled the number of filters in each layers. The topology of the best architectures can be found in the appendix for random search (A.1.1), REINFORCE (A.1.2) and PPO (A.1.3).

|           | outfilters=24 | outfilters=48 |
|-----------|---------------|---------------|
| Random    | 19.37%        | 15.04%        |
| REINFORCE | 13.72%        | 10.72%        |
| PPO       | 10.02%        | 8.21%         |

**Table 3.2:** Test error of the retrained best architectures found in case of the three search methods. We retrained the models with an outfilter of 24 and 48.

### 3.1.4.2 Results

Consequently, we have seen that training the controller with Proximal Policy Optimization yielded better results than the previously used REINFORCE as Figure 3.1 shows. PPO not only started to converge faster, but it also reached higher accuracy during searching and also in the retrained test results. With the retrained architectures we validated that the bias introduced into the architecture by training the controller leads to actual performance increase, although the difference in the retrained performance itself was marginal.

|                     | test error |
|---------------------|------------|
| Original NAS with RL | 2.65%      |
| ENAS                | 2.89%      |
| Our PPO-based ENAS  | 8.21%      |

**Table 3.3:** Test error of the retrained best architectures found in case of the original NAS, ENAS and our proposed PPO-based method

As shown in Table 3.3, we could not outperform the results of the original authors. One of the main contributing factors to this is that they applied a wide variety of tricks to boost the performance of models during the retraining phase in the end. However, our main intention with this paper was not to outperform previous approaches in terms of accuracy, but to show that PPO is a more suitable algorithm for driving the process of reinforcement learning in the controller, because it is capable of learning from fewer samples, thus it is more sample-efficient.

# Chapter 4

# Summary

First of all, we briefly covered the theory behind neural architecture search with reinforcement learning and also discussed the relevant, already existing methods in the field. Then, we proposed to modify the learning of the controller in ENAS from REINFORCE to PPO, because it is an algorithm that has demonstrated to be more stable and sample-efficient than his predecessor in different problem domains. To prove our claims, we created our own implementation of neural architecture search with both REINFORCE- and PPO-driven controllers and tested them for image classification on CIFAR-10 using a macro search space. We briefly analysed our results, and found that the PPO controller slightly outperformed our implementation of REINFORCE in terms of validation accuracy and was definitely better in terms of sample-efficiency as it achieved a higher accuracy, while seeing the same number of samples throughout training.

## 4.1 Future Work

During our work, we discovered several potential directions for continuing research in the area of neural architecture search. First of all, we plan to further explore the robustness of the proposed method, by adding a separate critic network to the controller architecture. We also want to extend the search space to different problem domains, such as image segmentation or sequential modelling. The design of recurrent cell structures is an appealing problem to us and probably we will proceed in this direction. We also want to maintain the open-source repository that we released here and therefore contribute to the reproducibility of this domain.

# Acknowledgements

We would like to express our graditude to both of our supervisors, Hamdi Abed and Bálint Gyires-Tóth for continuously supporting us with useful advices through the course of this work.

# List of Figures

# Bibliography

[1] Efficient neural architecture search via parameter sharing. https://github.com/melodyguan/enas, 2018.

[2] Hamdi M Abed. Introduction to neural architecture search (reinforcement learning approach). URL https://medium.com/@SmartLabAI/introduction-to-neural-architecture-search-reinforcement-learning-approach-55604772f173. Accessed: 2019-10-22.

[3] Marcin Andrychowicz, Filip Wolski, Alex Ray, Jonas Schneider, Rachel Fong, Peter Welinder, Bob McGrew, Josh Tobin, OpenAI Pieter Abbeel, and Wojciech Zaremba. Hindsight experience replay. In *Advances in Neural Information Processing Systems*, pages 5048–5058, 2017.

[4] Mohammad Ashraf. Reinforcement learning demystified: Markov decision processes (part 1)). URL https://towardsdatascience.com/reinforcement-learning-demystified-markov-decision-processes-part-1-bf00dda41690. Accessed: 2019-10-26.

[5] Dzmitry Bahdanau, Kyunghyun Cho, and Yoshua Bengio. Neural machine translation by jointly learning to align and translate. *arXiv preprint arXiv:1409.0473*, 2014.

[6] Yoshua Bengio, Patrice Simard, Paolo Frasconi, et al. Learning long-term dependencies with gradient descent is difficult. *IEEE transactions on neural networks*, 5(2):157–166, 1994.

[7] Greg Brockman, Vicki Cheung, Ludwig Pettersson, Jonas Schneider, John Schulman, Jie Tang, and Wojciech Zaremba. Openai gym. *arXiv preprint arXiv:1606.01540*, 2016.

[8] François Chollet. Xception: Deep learning with depthwise separable convolutions. In *Proceedings of the IEEE conference on computer vision and pattern recognition*, pages 1251–1258, 2017.

[9] George Cybenko. Approximation by superpositions of a sigmoidal function. *Mathematics of control, signals and systems*, 2(4):303–314, 1989.

[10] Ian Goodfellow, Yoshua Bengio, and Aaron Courville. *Deep Learning*. MIT Press, 2016. http://www.deeplearningbook.org.

[11] Kaiming He, Xiangyu Zhang, Shaoqing Ren, and Jian Sun. Deep residual learning for image recognition. In *Proceedings of the IEEE conference on computer vision and pattern recognition*, pages 770–778, 2016.

[12] Sepp Hochreiter and Jürgen Schmidhuber. Long short-term memory. *Neural computation*, 9(8):1735–1780, 1997.

[13] Kurt Hornik, Maxwell Stinchcombe, and Halbert White. Multilayer feedforward networks are universal approximators. *Neural networks*, 2(5):359–366, 1989.

[14] Sergey Ioffe and Christian Szegedy. Batch normalization: Accelerating deep network training by reducing internal covariate shift. *arXiv preprint arXiv:1502.03167*, 2015.

[15] Alex Krizhevsky, Geoffrey Hinton, et al. Learning multiple layers of features from tiny images. Technical report, Citeseer, 2009.

[16] Yann LeCun, Patrick Haffner, Léon Bottou, and Yoshua Bengio. Object recognition with gradient-based learning. In *Shape, contour and grouping in computer vision*, pages 319–345. Springer, 1999.

[17] Yann LeCun, Yoshua Bengio, and Geoffrey Hinton. Deep learning. *nature*, 521(7553): 436, 2015.

[18] Marius Lindauer and Frank Hutter. Best practices for scientific research on neural architecture search. *arXiv preprint arXiv:1909.02453*, 2019.

[19] Hanxiao Liu, Karen Simonyan, and Yiming Yang. Darts: Differentiable architecture search. *arXiv preprint arXiv:1806.09055*, 2018.

[20] Ilya Loshchilov and Frank Hutter. Sgdr: Stochastic gradient descent with warm restarts. *arXiv preprint arXiv:1608.03983*, 2016.

[21] Mitchell Marcus, Beatrice Santorini, and Mary Ann Marcinkiewicz. Building a large annotated corpus of english: The penn treebank. 1993.

[22] Adil Moujahid. A practical introduction to deep learning with caffe and python. URL http://adilmoujahid.com/posts/2016/06/introduction-deep-learning-python-caffe/. Accessed: 2019-10-25.

[23] Vinod Nair and Geoffrey E Hinton. Rectified linear units improve restricted boltzmann machines. In *Proceedings of the 27th international conference on machine learning (ICML-10)*, pages 807–814, 2010.

[24] Christopher Olah. Understanding lstm networks. URL https://colah.github.io/posts/2015-08-Understanding-LSTMs. Accessed: 2019-10-17.

[25] Hieu Pham, Melody Y Guan, Barret Zoph, Quoc V Le, and Jeff Dean. Efficient neural architecture search via parameter sharing. *arXiv preprint arXiv:1802.03268*, 2018.

[26] David E Rumelhart, Geoffrey E Hinton, Ronald J Williams, et al. Learning representations by back-propagating errors. *Cognitive modeling*, 5(3):1, 1988.

[27] Tom Schaul, John Quan, Ioannis Antonoglou, and David Silver. Prioritized experience replay. *arXiv preprint arXiv:1511.05952*, 2015.

[28] John Schulman, Sergey Levine, Pieter Abbeel, Michael Jordan, and Philipp Moritz. Trust region policy optimization. In *International conference on machine learning*, pages 1889–1897, 2015.

[29] John Schulman, Philipp Moritz, Sergey Levine, Michael Jordan, and Pieter Abbeel. High-dimensional continuous control using generalized advantage estimation. *arXiv preprint arXiv:1506.02438*, 2015.

[30] John Schulman, Filip Wolski, Prafulla Dhariwal, Alec Radford, and Oleg Klimov. Proximal policy optimization algorithms. *arXiv preprint arXiv:1707.06347*, 2017.

[31] Kenneth O Stanley and Risto Miikkulainen. Efficient reinforcement learning through evolving neural network topologies. In *Proceedings of the 4th Annual Conference on Genetic and Evolutionary Computation*, pages 569–577. Morgan Kaufmann Publishers Inc., 2002.

[32] Richard S. Sutton and Andrew G. Barto. *Reinforcement Learning: An Introduction*. MIT Press, 1998. URL http://www.cs.ualberta.ca/~sutton/book/the-book.html.

[33] Richard S Sutton, David A McAllester, Satinder P Singh, and Yishay Mansour. Policy gradient methods for reinforcement learning with function approximation. In *Advances in neural information processing systems*, pages 1057–1063, 2000.

[34] Christian Szegedy, Wei Liu, Yangqing Jia, Pierre Sermanet, Scott Reed, Dragomir Anguelov, Dumitru Erhan, Vincent Vanhoucke, and Andrew Rabinovich. Going deeper with convolutions. In *Proceedings of the IEEE conference on computer vision and pattern recognition*, pages 1–9, 2015.

[35] Sik-Ho Tsang. Review: Mobilenetv1 — depthwise separable convolution (light weight model). URL https://towardsdatascience.com/review-mobilenetv1-depthwise-separable-convolution-light-weight-model-a382df364b69. Accessed: 2019-10-25.

[36] Ziyu Wang, Victor Bapst, Nicolas Heess, Volodymyr Mnih, Remi Munos, Koray Kavukcuoglu, and Nando de Freitas. Sample efficient actor-critic with experience replay. *arXiv preprint arXiv:1611.01224*, 2016.

[37] Paul J Werbos et al. Backpropagation through time: what it does and how to do it. *Proceedings of the IEEE*, 78(10):1550–1560, 1990.

[38] Ronald J Williams. Simple statistical gradient-following algorithms for connectionist reinforcement learning. *Machine learning*, 8(3-4):229–256, 1992.

[39] Barret Zoph and Quoc V Le. Neural architecture search with reinforcement learning. *arXiv preprint arXiv:1611.01578*, 2016.

# Appendix

## A.1 Topology of the best architectures found during the architecture searches

The elliptic blocks and the dotted arrows denote blocks and skip connections that were found during the search, while rectangular elements and normal lines are fixed components.

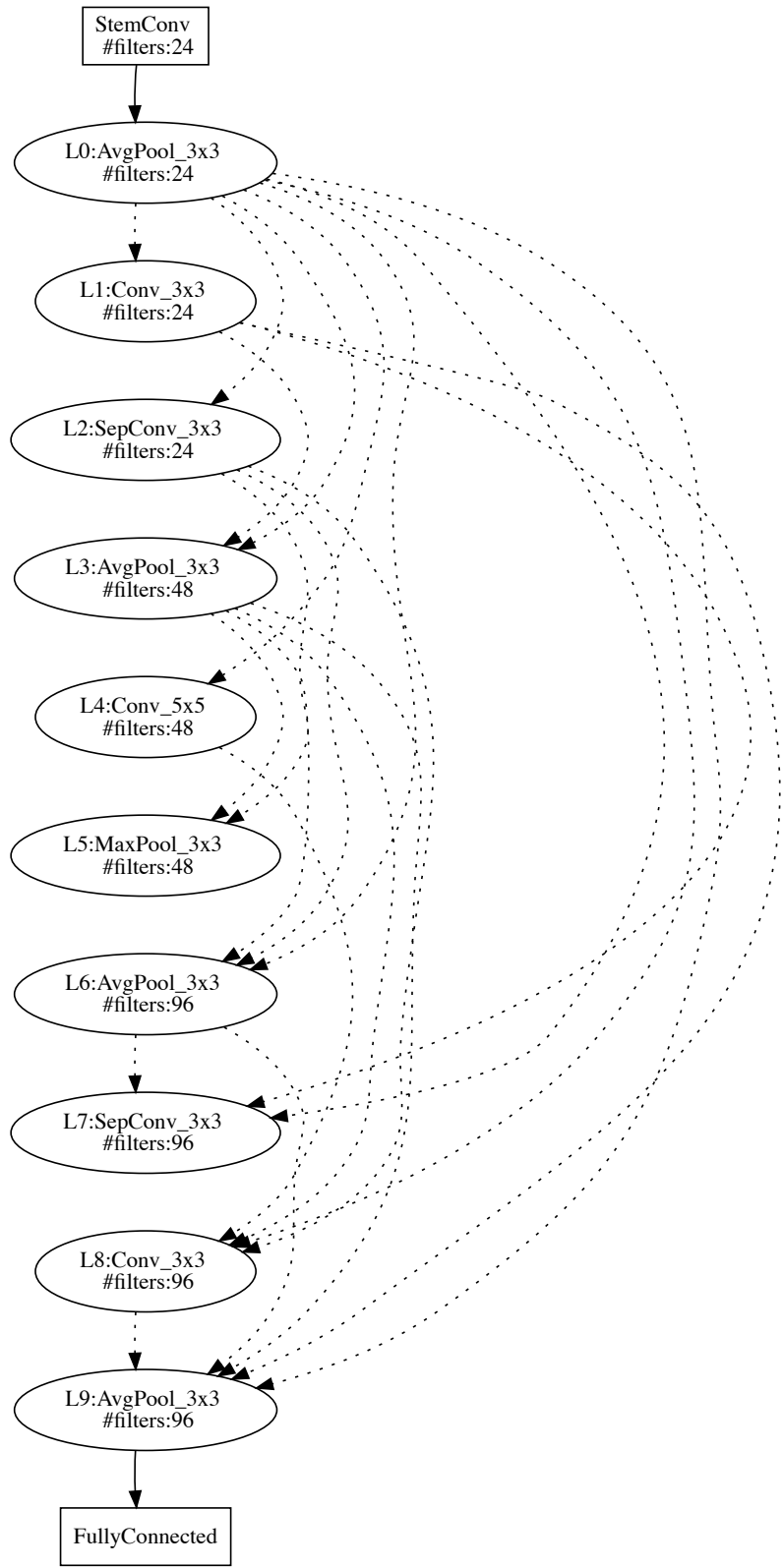

**Figure A.1.1:** The best-performing architecture found by the random search

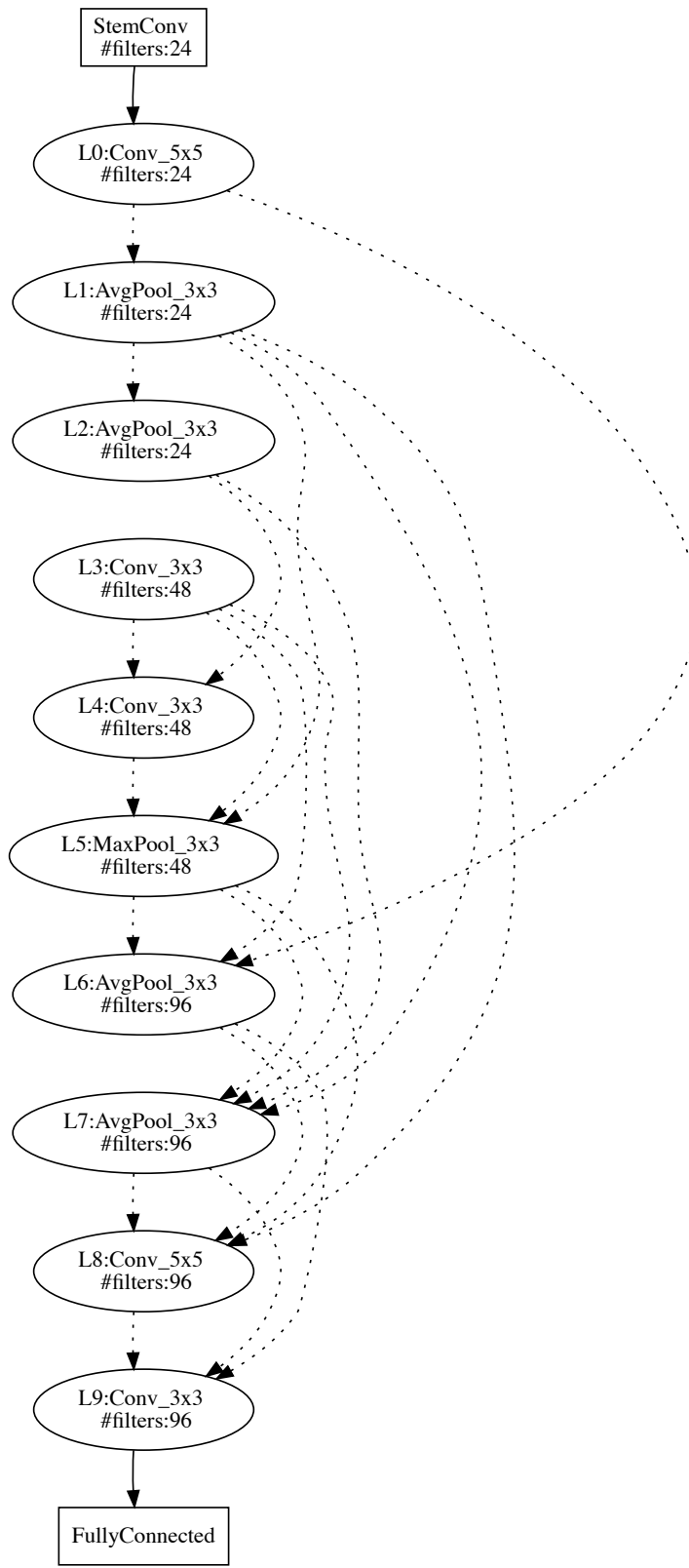

**Figure A.1.2:** The best-performing architecture found by the RE-INFORCE algorithm

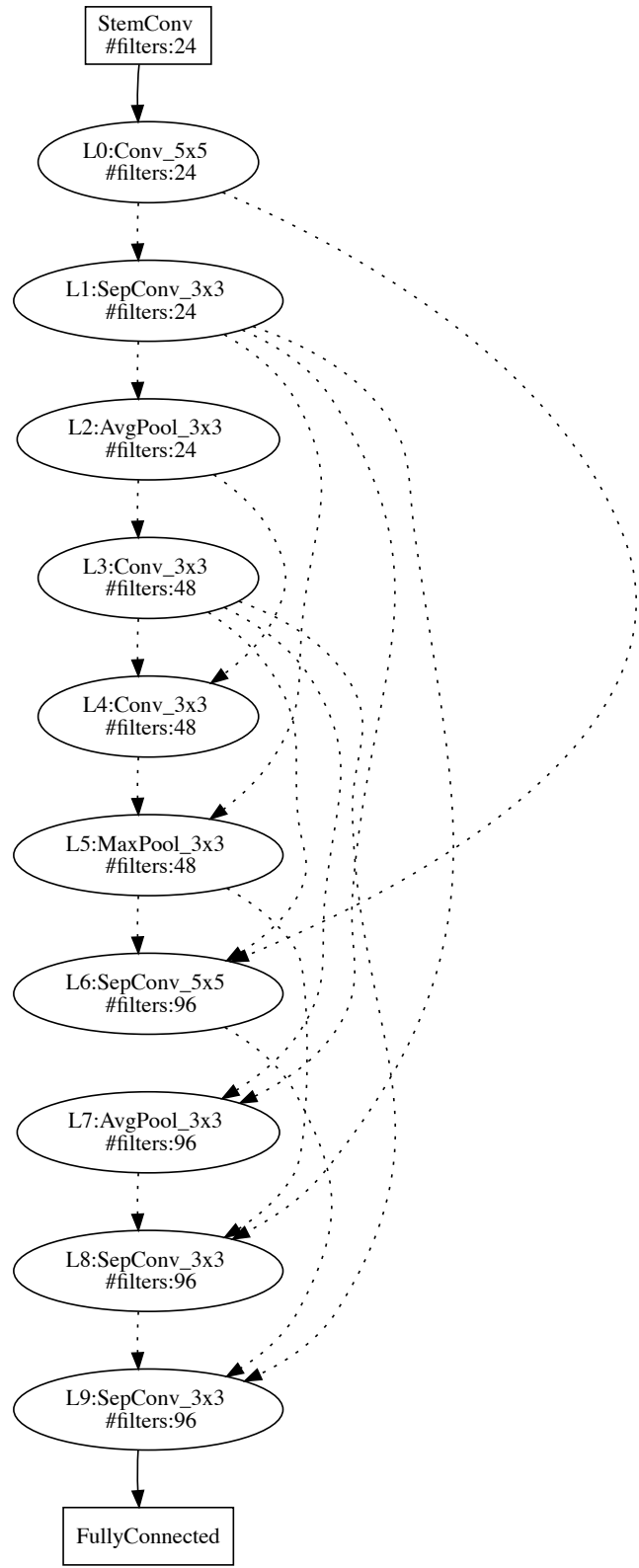

**Figure A.1.3:** The best-performing architecture found by our proposed PPO-based ENAS